\documentclass[11pt,a4paper]{article}
\usepackage[hyperref]{acl2020}
\usepackage{times}
\usepackage{latexsym}

\usepackage{microtype}

\usepackage[normalem]{ulem}
\useunder{\uline}{\ul}{}
\usepackage{graphicx}
\usepackage{booktabs,tabularx}
\usepackage{multirow}
\usepackage{tabularx}

\aclfinalcopy 

\title{How toxic is antisemitism? Potentials and limitations of automated toxicity scoring for antisemitic online content}
  \author{Helena Mihaljević \\
  HTW Berlin \\ 
  \texttt{mihalje@htw-berlin.de} \\\And
  Elisabeth Steffen \\
  HTW Berlin\\ 
  \texttt{steffen@htw-berlin.de} \\}

\date{}

\begin{document}
\maketitle
\begin{abstract}

The Perspective API, a popular text toxicity assessment service by Google and Jigsaw, has found wide adoption in several application areas, notably content moderation, monitoring, and social media research.
We examine its potentials and limitations for the detection of antisemitic online content that, by definition, falls under the toxicity umbrella term. Using a manually annotated German-language dataset comprising around 3,600 posts from Telegram and Twitter, we explore as how toxic antisemitic texts are rated and how the toxicity scores differ regarding different subforms of antisemitism and the stance expressed in the texts. We show that, on a basic level, Perspective API recognizes antisemitic content as toxic, but shows critical weaknesses with respect to non-explicit forms of antisemitism and texts taking a critical stance towards it. Furthermore, using simple text manipulations, we demonstrate that the use of widespread antisemitic codes can substantially reduce API scores, making it rather easy to bypass content moderation based on the service's results.
\end{abstract}

\section{Introduction}

The current COVID-19 pandemic has been accompanied by an increase in insults, hostility, and hate speech, often directed at Jews who (once again) have been singled out as one of the main culprits in times of crises. A recent large scale study of multiple online platforms reveals that ``almost 35\% of all posts mentioning Jews or Jewishness expressed negativity toward Jews'', with toxic speech against Jews amounting for two to five percent of the posts in some forums \cite{cohen_antisemitism_2021}. With regard to the current `infodemic', antisemitism is of special relevance as it shares relevant features and is often deeply intertwined with conspiracy theories: 
Both are based on simplifying forms of personification in combination with a Manichean worldview and the ontological construction of group identities \cite{haury_antisemitismus_2002}. The hostility towards Jews and other targeted groups expressed in these narratives has a negative impact not only on digital spheres but also reaches out to the real world, amplifying verbal and physical acts of violence. It is thus of great importance for a variety of stakeholders such as content moderators, researchers or NGOs monitoring societal developments to have access to tools for automated detection of antisemitic online content. 

Despite the current rise of antisemitic conspiracy theories, and the hateful, toxic characteristics of antisemitism, the phenomenon is still under-explored in large-scale research of online content in general, and hate-speech in particular \cite{steffen_codes_2022}. To the best of our knowledge, there are currently no services for automated detection of antisemitic content. However, progress has been made in form of datasets, code packages, and production-ready web services regarding the recognition of other linguistic phenomena intersecting with antisemitism, such as hate speech and toxic language. Perspective API, a free service created by Jigsaw and Google's Counter Abuse Technology team, is one such widely used technology. It allows the detection of abusive content by computing scores for different attributes such as toxicity, insult or threat. Perspective API could thus provide a low-threshold approach to detect certain forms of antisemitic speech and include it in monitoring and moderation efforts. This paper aims to explore the possibilities of the service with regard to this objective.

Accordingly, we address the following research questions:

\begin{itemize}
\itemsep-0.5em 
\item RQ1: As how toxic are antisemitic texts rated?
\item RQ2: Are encoded manifestations of antisemitism rated as toxic?
\item RQ3: Are critiques of antisemitic statements rated as toxic?
\item RQ4: How do modifications of antisemitic statements affect their toxicity score?
\end{itemize}

We use a set of  3.642 German-language Telegram and Twitter posts published during the COVID-19 pandemic, annotated in terms of content and stance with respect to antisemitism. We evaluate different attributes of the Perspective API that, by definition, should produce higher scores when confronted with antisemitic texts. We further analyze the scores depending on the subform of antisemitism and the stance towards it. Finally, we perform adversary attacks to assess in how far modifications of antisemitic statements influence their scores. 

\paragraph{Content Warning}
This article contains examples of hateful content including offensive, insulting and threatening comments targeting Jewish people but also other individuals and groups frequently targeted by antisemitic hate speech. It might therefore cause anxiety among members of various population groups.

\section{Related Work}
\label{sec:related_work}

Similar to other sociolinguistic phenomena such as offensive or abusive language, toxicity is not uniquely defined across existing research and rather used as an umbrella term. \citet{horta_ribeiro_platform_2021} refer to it as ``socially undesirable content'' that includes ``sexist, racist, homophobic, or transphobic posts, targeted harassment, and conspiracy theories that target racial or political groups''. 
In \citet{cohen_antisemitism_2021} it is understood as ``blatantly aggressive and demeaning messages about a group or person, such as dehumanization, incitement of hatred or discrimination, or justification of violence''. The authors note that ``toxic language includes but is not limited to hate speech'', but in fact utilize machine learning models developed for hate speech detection. 

Perspective API defines ``rude, disrespectful or unreasonable'' content that is ``likely to make people leave a discussion'' as toxic \cite{google_about_2022,thain_wikipedia_2017} and provides scores representing the likelihood that a reader will perceive a text as e.g. toxic.
The service is used in a variety of applications such as The New York Times website and the social news platform Reddit \cite{google_about_2022}, while also being applied by social and online media scholars. It has been used as a pre-filtering method for analyses of moderation measures on Reddit \cite{horta_ribeiro_platform_2021}, investigations of political online communities \cite{rajadesingan_quick_2020} such as the QAnon movement 
\cite{hoseini_globalization_2021}, or to identify antisemitic and islamophobic texts on 4chan that are subsequently used for the detection of hateful images via contrastive learning \cite{gonzalez-pizarro_understanding_2022}.

Perspective API also measures severe toxicity of a text as, roughly speaking, an even stronger form of toxicity. 
A severe toxicity score of 0.8 is chosen as the lower limit in a number of studies to preselect particularly toxic texts \cite{horta_ribeiro_platform_2021,hoseini_globalization_2021,rajadesingan_quick_2020,zannettou_measuring_2020}. The toxicity scores provided by Perspective API have been validated on random manually labeled text samples in e.g. \citet{horta_ribeiro_platform_2021} and \citet{gehman_realtoxicityprompts_2020}. \citet{horta_ribeiro_platform_2021} compared its toxicity scores with results from HateSonar, a tool developed for the detection of hate speech and offensive language \cite{davidson_automated_2017}, deducing that Perspective API yields better results (however, the evaluation is based on a rather small sample of data). 

Despite its broadness and ambiguity, in the definition of Perspective API and beyond, the term toxicity encompasses antisemitic speech with its widely accepted operational definition as ``a certain perception of Jews, which may be expressed as hatred toward Jews'' \cite{international_holocaust_remembrance_alliance_working_2016}.
However, the design of the service already indicates the possibility of certain shortcomings with respect to detecting toxic antisemitic texts. As the developers of the Perspective API themselves point out, the very definition of toxic language has a subjective character  \cite{borkan_nuanced_2019}. 
Some labeled datasets used for training the respective models were published as part of Kaggle competitions to improve models and reduce unintended model bias \cite{thain_wikipedia_2017,wulczyn_ex_2017,borkan_nuanced_2019}. Labeling by a larger number of crowdworkers is used as a vehicle to make the dataset and the models trained on it more robust. However, a look at the annotated data exposes various examples confirming that this is not sufficient to label antisemitic content as toxic. For instance, the following text was annotated by 54 crowdworkers, with an average toxicity score of 0.33 \cite{thain_wikipedia_2017}: ``The US has finally cut bait on the occultist blood suckers. Obama and Trump just drop kicked bibi down to size. This has been a long time coming and that is why the zionists wanted Hitlery to win and start ww3.''
Research has shown that the annotation of antisemitic content poses considerable difficulties, 
even for scholars with respective backgrounds \cite{ozalp_antisemitism_2020,steffen_codes_2022}, thus it is plausible that annotators assess antisemitic texts in diverging ways. The task is further complicated by the fact that antisemitism is often expressed implicitly, using codes which annotators need to be familiar with \cite{jikeli_annotating_2019} or additional context \cite{jikeli_toward_2022} in order to recognize them as antisemitic language. Furthermore, toxicity can be significantly lowered by undertaking minor changes such as single character-level insertions or perturbations in words associated with toxicity (e.g. `stupid' $\rightarrow$ `st.upid'), while the scores remain relatively high, if the statement is negated \cite{hosseini_deceiving_2017}.

The service has also been shown to be biased with respect to differences in dialect, computing a significantly higher score for texts in African American English \cite{sap_risk_2019}. Recent work demonstrated that systems tend to produce false positive bias by overestimating the level of toxicity if minorities are mentioned \cite{dixon_measuring_2018,hutchinson_social_2020}. \citet{rottger_hatecheck_2021} developed functional tests for hate speech detection models and evaluated the Perspective API and three other models. Their results indicate that all models have critical weaknesses, namely an over-sensitivity to certain keywords, a common misclassification of non-hateful content (such as counterspeech), and statements including reclaimed slurs. Furthermore, the models were biased across the different target groups included in the test data (women, trans people, gay people, Black people, disabled people, Muslims, and immigrants; Jews were not included).

\section{Antisemitic language}

As a basic definition, we apply the working definition by the International Holocaust Remembrance Alliance of antisemitism as ``a certain perception of Jews, which may be expressed as hatred toward Jews'' which can be ``directed toward Jewish or non-Jewish individuals and/or their property, toward Jewish community institutions and religious facilities'' \cite{international_holocaust_remembrance_alliance_working_2016}.
Among several other narrative strategies this may include calls for the killing or harming of Jews as well as false, dehumanizing, demonizing, or stereotyping accusations against Jews or the power of Jews as a collective. 

We extend the working definition to also include certain subforms of antisemitism we consider as specifically relevant in the context of the COVID-19 pandemic, namely \textit{encoded antisemitism} and \textit{post-Holocaust antisemitism}. 

Encoded forms of antisemitism are statements which do not mention Jews or the State of Israel, but instead turn generally against presumed or actual economic or political elites while deploying antisemitic codes or stereotypes,  e.g. narratives holding Bill Gates or `Big Pharma' accountable for inventing and/or benefiting from the COVID-19 pandemic and the resulting global crisis instead of explicitly mentioning and accusing Jews as initiators. 
Common stereotypes which are by no means exhaustive are:  Actually or allegedly Jewish persons or dynasties such as Rothschild, Rockefeller, George Soros, Mark Zuckerberg, or Bill Gates; animal metaphors: e.g. octopus, snake, pig, rat; disease and cancer metaphors such as virus, germ, parasite, cancer; codes referring to the `lying press' trope (`Lügenpresse', `Pinocchio-Presse', `Systemmedien'); and codes referring to a financial (Jewish) elite in control of global events ('financial elite', 'high finance', 'East coast', 'Wall street'). Note that the occurrence of a single code 
is typically not sufficient to label a text as antisemitic and that antisemitic codes can be articulated consciously as well as unconsciously.
 
Manifestations of post-Holocaust antisemitism explicitly name Jews as part of argumentation strategies which instrumentalize the victims of the Holocaust for a political agenda and at the same time shift the perpetrator-victim coordinates by undertaking relativizing Holocaust comparisons. In the context of the COVID-19 pandemic, we encounter forms of post-Holocaust antisemitism in comparisons or equations of the state measures against the pandemic with the Nazi persecution of Jews. Common examples are the use of the term `Giftspritze' for COVID-19 vaccinations as a more or less implicit reference to the illegal and often lethal experiments performed on human beings by the Nazis, the use of the yellow star with the imprint `Ungeimpft' (unvaccinated) by which anti-vaccination protesters compare themselves to Jews under the Nazi regime, and references to known victims of and/or resistance fighters against the Nazi regime such as Anne Frank or Sophie Scholl. 

\section{Data and methods}
Our data was annotated using a comprehensive annotation scheme developed as part of a research project on online antisemitism and conspiracy narratives in the context of the COVID-19 pandemic. The scheme consists of two main categories, antisemitism and conspiracy theory, and sub-labels to specify the content and stance of a message. The scheme and the annotated dataset are described in detail in \citet{steffen_codes_2022}.\footnote{The cited manuscript has been submitted for publication, thus the dataset and the annotation scheme are not yet publicly available. Until the publication, all documents and data can be made available to researchers upon request.}

The annotation was performed by a team of nine researches with scientific backgrounds in political science, sociology, or data science. We annotated a corpus consisting of a few thousand messages from Telegram and Twitter. While most of the messages were labeled by a single individual, the annotation process was continuously reflected in regular discussions and a joint workshop. We furthermore evaluated inter-annotator reliability on an additional sample of 445\footnote{Of the 500 texts originally selected at random, 55 were excluded because, for example, they were not in German, were too short, or were incomprehensible.} records, yielding Cohen's kappa of $\kappa=0.84$ and thus strong agreement for the category antisemitism.
  
For the experiments on the Perspective API presented here, we use a dataset consisting of 3.642 texts, with $\sim\!\!3.200$ Telegram messages and $\sim\!400$ Twitter tweets. Around $19\%$ of all posts were classified as articulating and/or addressing antisemitism, with $\sim\!40.6\%$ labeled as encoded antisemitism, $\sim\!29\%$ as Post-Holocaust antisemitism, and $\sim\!29\%$ as explicit forms of antisemitism. The stance expressed was predominantly affirmative ($\sim\!68\%$), while $\sim\!24.4\%$ of the texts expressed a critical stance ($\sim\!24.4\%$), and the fewest were classified as neutral or uncertain ($\sim\! 7.5\%$).

The two sources differ not only in terms of size but also regarding the sampling approaches: While we selected tweets from a dataset about the German `Querdenken' movement based on antisemitism-related keywords, the Telegram messages were sampled from pre-selected channels disseminating conspiracy theory content as well as critique of the anti-COVID-19 measures in Germany. It is thus not surprising that the Telegram dataset contains a greater variety of topics and thus a smaller proportion of antisemitic content ($\sim\! 14\%$) than the analyzed tweets ($\sim\! 56\%$)\footnote{Against this background, we believe that our results should not lead to the conclusion that antisemitism is generally more prevalent on Twitter than on Telegram.}.

\begin{figure}[h!]
\begin{center}
\includegraphics[width=\linewidth]{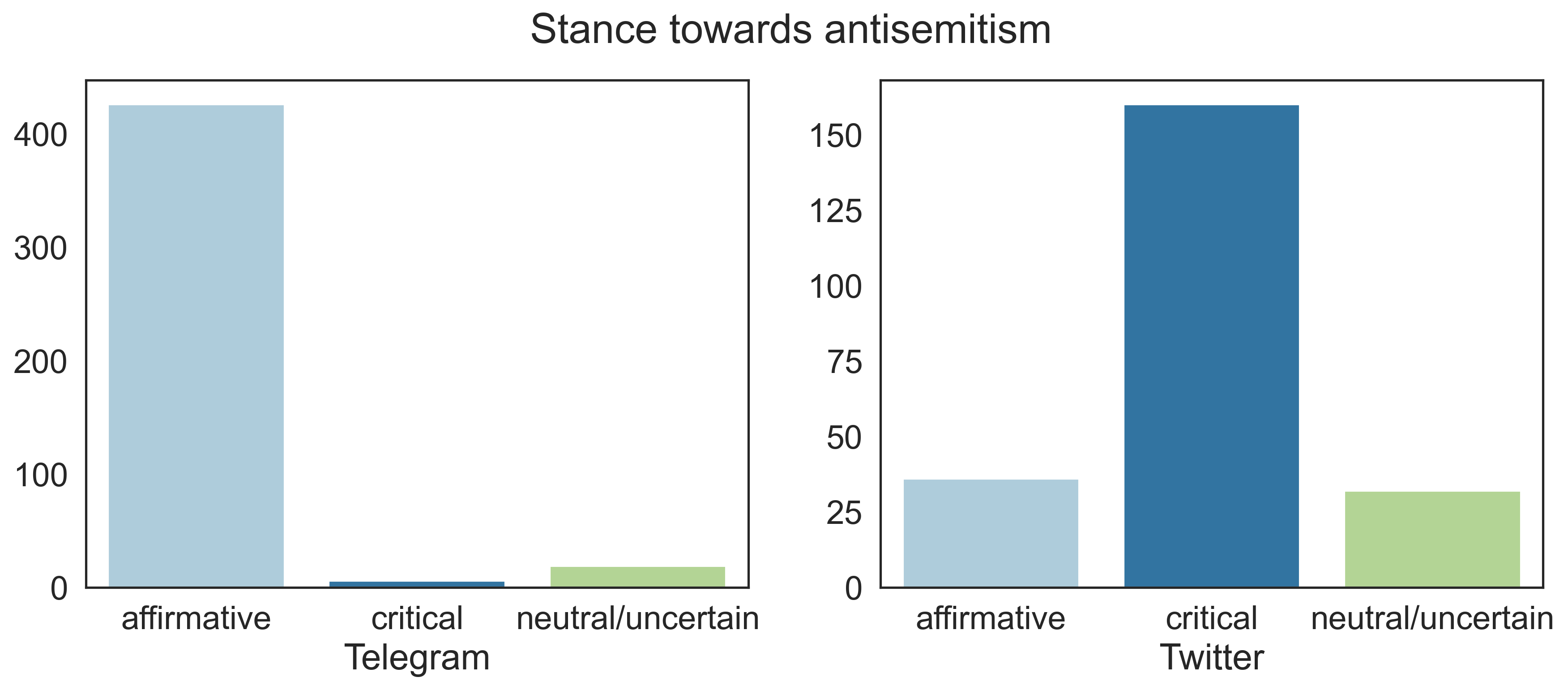}%
\end{center}
\caption{Stance towards antisemitic content: while only $\sim\! 16\%$ of tweets are classified as affirmative and over $\sim\! 70\%$  as critical, almost $95\%$ of all Telegram messages with antisemitic content were classified as affirmative.}
\label{fig:1}
\end{figure}

At the same time, almost all antisemitism-related texts from Telegram are affirmative towards antisemitism, while Twitter users in our dataset talk about antisemitism, but not necessarily support antisemitic worldviews (cf. Figure  \ref{fig:1}).

Encoded antisemitism makes up for almost half of the antisemitic content in our Telegram dataset ($\sim\!\!49\%$), followed by explicitly articulated antisemitism ($\sim\!\!37\%$), and post-Holocaust antisemitism ($\sim\!\!13\%$). On Twitter, post-Holocaust is clearly the dominant subform with $60\%$, followed by encoded ($\sim\!\!23\%$) and explicit antisemitism ($17\%$). We suppose that the prevalence of post-Holocaust antisemitism in our Twitter data is due to the fact that users critically addressed the German `Querdenken' movement and its comparisons of anti-COVID-19 regulations with the Nazi regime. 

We have retrieved the scores for the attributes \textit{insult}, \textit{identity attack}, \textit{threat}, \textit{toxicity} and \textit{severe toxicity} from Perspective API, since their definition \cite{google_about_2022} shares relevant features with antisemitic language as defined above (cf. Table 1 in Appendix). 
The returned score is a value between 0 and 1 that ``indicates how likely it is that a reader would perceive the comment provided in the request as containing the given attribute'' \cite{google_about_2022-1}.  

\section{Results}

Our results indicate that our dataset has a strong toxic bias: A median severe toxicity score of $\sim\!\!0.18$ clearly exceeds not only the baseline Telegram dataset compiled in \citet{hoseini_globalization_2021} with a median severe toxicity score of 0.03 but also their QAnon Telegram dataset (median: 0.07). The CDF further reveals that only $\sim\!20\%$ of the texts are assigned a severe toxicity score lower than 0.1, while this holds for around $60\%$ of all texts in the mentioned baseline set. While the Telegram subset is almost identical to all texts regarding its CDF and its median of 0.18, the Twitter subset is more toxic with a median of 0.29. Overall, the median scores range between $\sim\!\!0.18$ for severe toxicity, and $\sim\!\!0.35$ for insult and identity attack.

\subsection{Antisemitic content}
\begin{figure}[h!]
\begin{center}
\includegraphics[width=\linewidth]{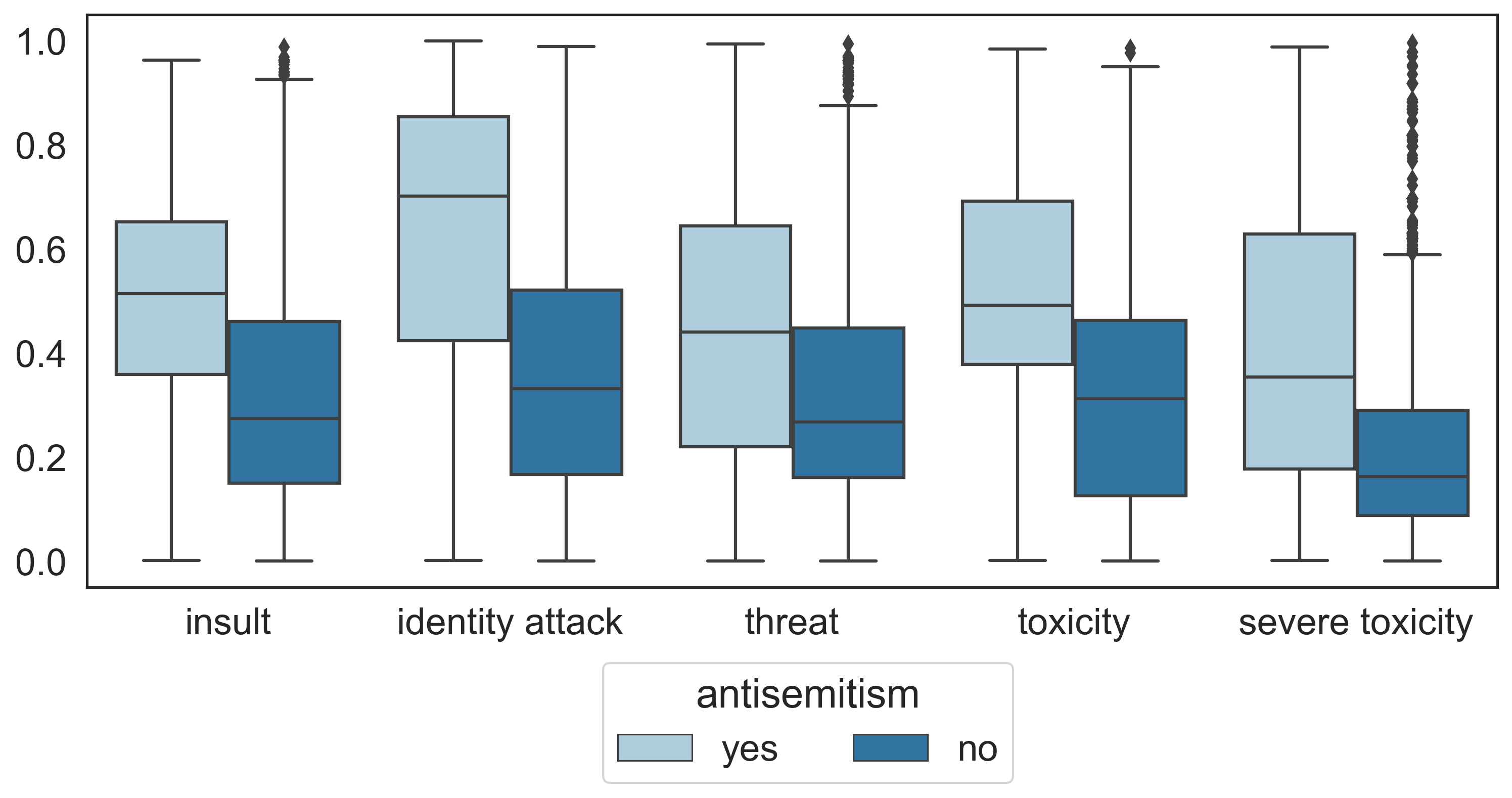}
\end{center}
\caption{Distribution of scores in relation to the presence of antisemitic content.}
\label{fig:2}
\end{figure}

Texts classified as antisemitic have higher scores than those not classified as such with respect to all attributes. As shown in Figure \ref{fig:2}, the median scores are around twice as high for texts containing antisemitism, with the greatest difference for severe toxicity (0.16 versus 0.35) and identity attack (0.33 versus 0.7). These findings support our hypothesis that texts with antisemitic content share relevant features with messages classified as threatening, toxic, etc. by the Perspective API. Furthermore, these types of texts often contain other insults and threats, which further contributes to the increase of their scores. 

\subsection{Subforms of antisemitism}

Explicit forms of antisemitism rank highest in all categories with medians between 0.56 and 0.83 (identity attack), while those for encoded forms of antisemitism range between 0.29 and 0.53. This supports our hypothesis that Perspective API generally does recognize antisemitic content as toxic, but finds it more difficult to recognize rather implicit forms of antisemitism. Texts communicating (about) narratives related to post-Holocaust antisemitism are ranked very similar to those classified as encoded antisemitism except for identity attack and severe toxicity, where they yield significantly higher values (0.53 vs. 0.63 and 0.29 vs. 0.35). The higher score, in particular for identity attack, might originate from more frequent mentions of the Holocaust and Nazis, which might also explain why there are no texts with a score of 0 in this category. The fact that most of these texts criticize the antisemitism of the `Querdenken' movement indicates that these two endpoints 
do not perform well on capturing the stance of the messages. 

\begin{figure}[h!]
\begin{center}
\includegraphics[width=\linewidth]{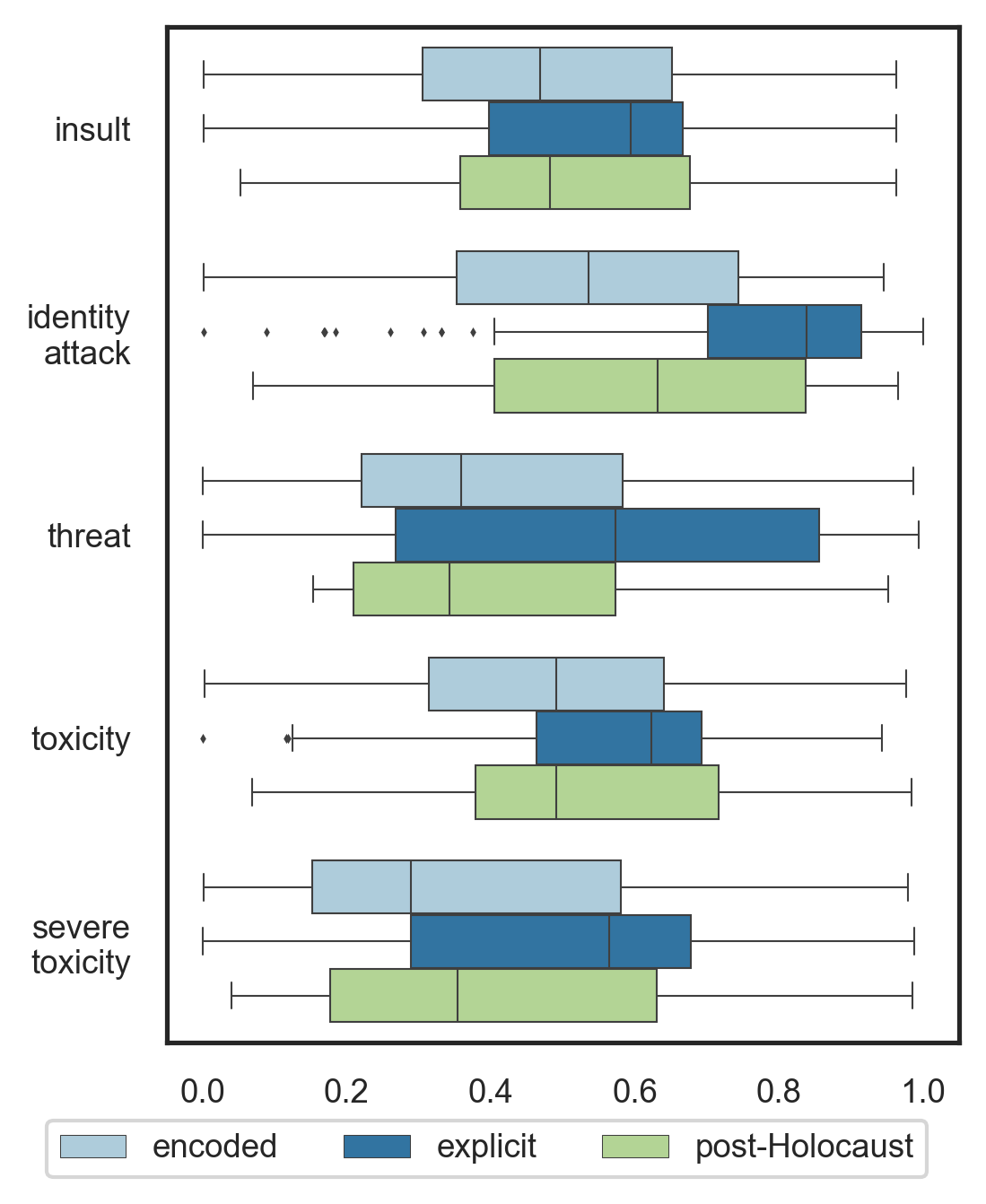}%
\end{center}
\caption{Distribution of scores in relation to the subform of antisemitism, with significantly higher values for explicitly antisemitic texts,}
\label{fig:3}
\end{figure}

\subsection{Stance towards antisemitism}

As indicated in the previous exploration, Perspective API might not be able to differentiate the stance of texts even at a basic level. We thus performed experiments with single sentences and phrases extracted from our dataset that address a similar topic but with different stance. One such example is presented in Table \ref{tab:1}. Note that the critical example does not contain any other kinds of toxic language or hate speech, thus side effects can be ruled out. The scores for the critical example clearly outnumber the affirmative text in all categories. Furthermore, all scores are in a remarkably high range with 
$\sim\!0.7$ for insult and $> 0.9$  for identity attack. By contrast, even the highest 
score for the affirmative example is lower than $\sim\!0.2$, and in most of the categories close to 0. 

\begin{table*}[]
\begin{tabular}{p{0.5\textwidth} | p{0.07\textwidth} | p{0.07\textwidth} | p{0.07\textwidth} | p{0.07\textwidth} | p{0.08\textwidth} } 
text & insult & identity attack & threat & toxicity & severe toxicity \\
\hline
The Holocaust was unique in its contempt for humanity and its consequences for the world community.\footnote{German original: Der Holocaust war einzigartig in seiner Menschenverachtung und den Folgen für die Weltgemeinschaft.} & 0.67 & 0.93 & 0.85  & 0.79 & 0.89 \\
The Holocaust did not happen.\footnote{German original: Der Holocaust hat nicht stattgefunden.}  & 0.05  & 0.07  & 0.15 & 0.13 & 0.04                          
\end{tabular}
\caption{Expression of the historical significance of the Holocaust is assessed as particularly toxic, while its negation is rated with very low scores.}
\label{tab:1}
\end{table*}

\begin{figure}[h!]
\begin{center}
\includegraphics[width=\linewidth]{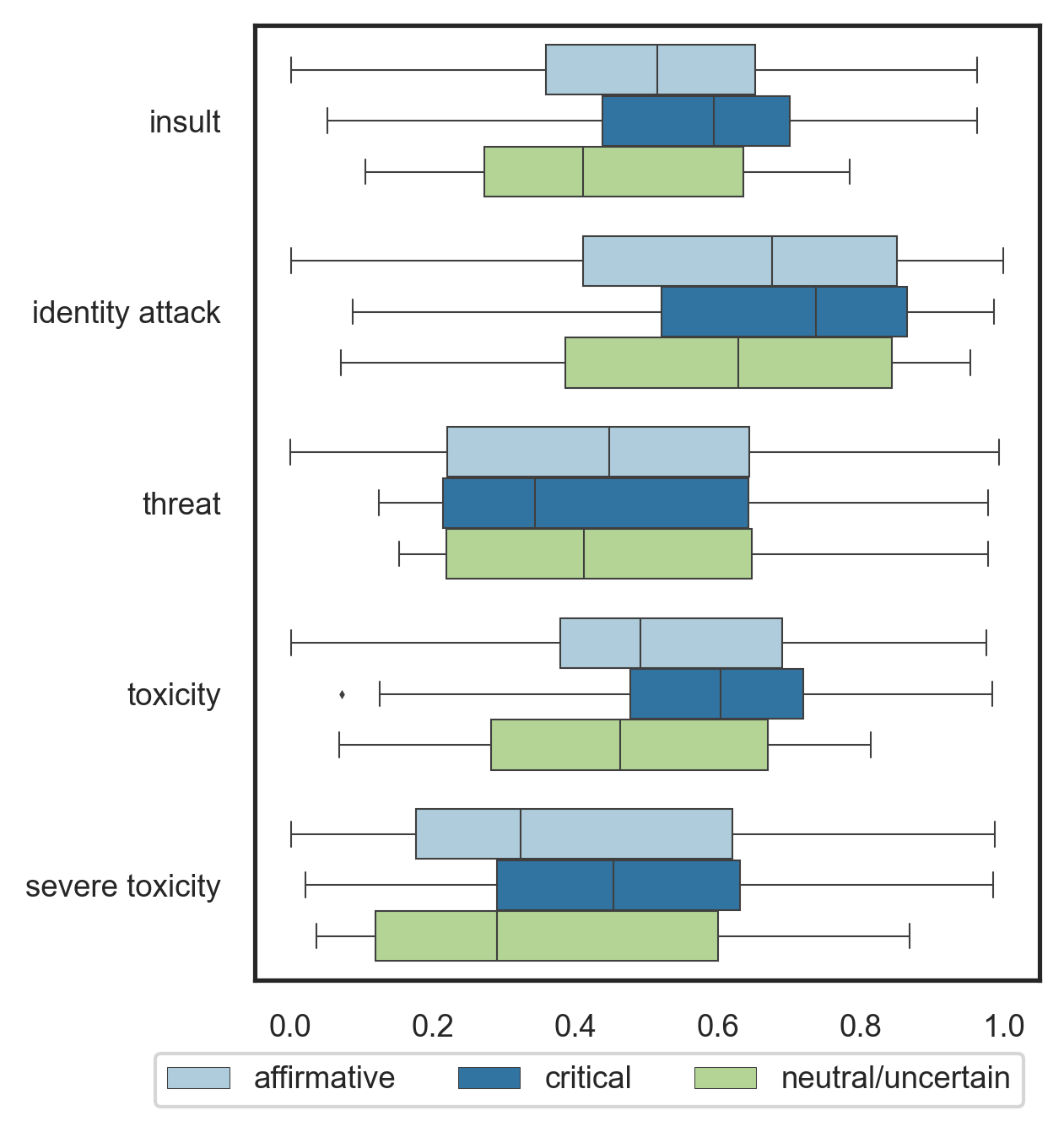}%
\end{center}
\caption{Texts taking a critical stance are ranked highest in all categories except threat, with medians between 0.45 and 0.74 (threat: 0.34). }
\label{fig:4}
\end{figure}

As shown in Figure \ref{fig:4}, our statistical exploration confirms the observed tendency for the entire dataset, with texts taking a critical stance towards antisemitism ranked highest in almost all categories. Texts with affirmative or neutral/uncertain stance towards antisemitism are ranked roughly equally. 

\subsection{Adversarial attacks: the impact of text manipulations}
We are interested in how the Perspective API reacts to explicit mentions of Jews or Israel in comparison to antisemitic codes. To analyze this, we conducted two experiments: 

First, we added `\#Israel' and `\#Juden' (`\#Jews') to all texts in the corpus labeled as antisemitic ($n=679$). Since prior work indicates an oversensitivity of hate speech detection models to certain keywords \cite{rottger_hatecheck_2021}, our aim was to evaluate if the presence of words explicitly related to Jewishness would affect the API's assessment. 

Our second experiment focussed on antisemitic codes frequently observed in online content \cite{zannettou_measuring_2020,finkelstein_antisemitic_2020}. Such codes allow to express antisemitic worldview without explicitly expressing hatred against Jewish individuals or communities, thus avoiding social ostracism as well as platform bans or criminal prosecution. We believe that they can play a crucial role for spreading antisemitism, since their subtlety hampers their detection and facilitates their dissemination. To evaluate the API's assessment of encoded antisemitism, we replaced words related to Jews or Jewishness by frequently observed antisemitic codes, namely `zionist', `globalist' and `satanist Freemason'\footnote{German words: `Zionist', `Globalist', `satanistischer Freimaurer'}, in texts labeled as explicitly and affirmatively antisemitic ($n=89$). Examples of replacements are listed in Table 2 in the Appendix. In addition, we used triple parentheses, a ``widely used antisemitic symbol that calls attention to supposed secret Jewish involvement and conspiracy'' \cite{zannettou_measuring_2020}, known from online communication in English-language context.

We focus on the attributes identity attack, toxicity and severe toxicity since we expect these to be affected most by the performed manipulations. 

\begin{figure}[h!]
\begin{center}
\includegraphics[width=\linewidth]{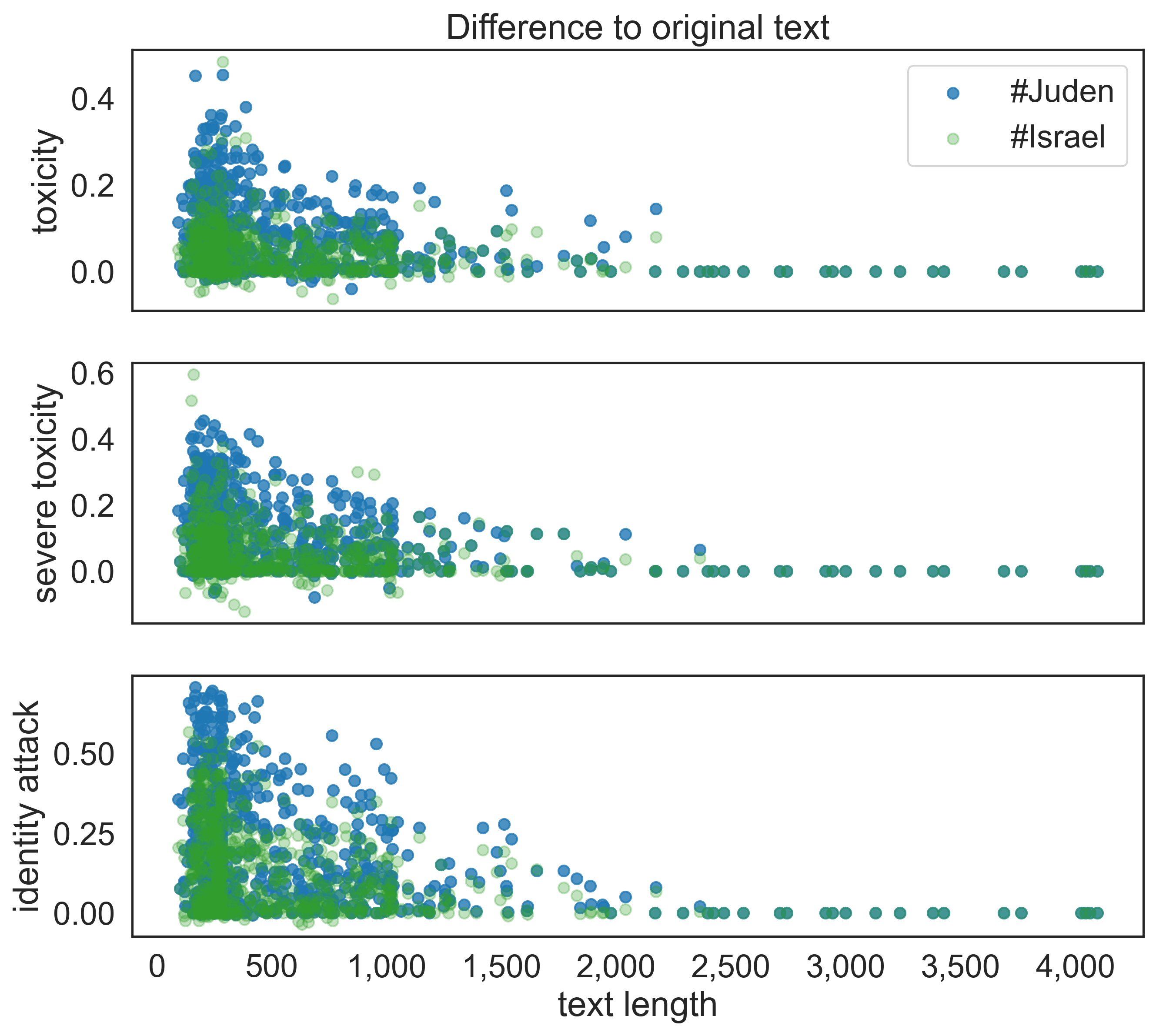}%
\end{center}
\caption{Difference between scores of manipulated and original message, where manipulations consist of adding one of the two string `\#Israel' and `\#Juden' to the end of the text ($n=679$).}
\label{fig:5}
\end{figure}

Figure \ref{fig:5} shows the difference between the scores achieved when adding either `\#Israel' or `\#Juden' to the end of a text and a text's original score, in relation to the length of a text. Clearly, the effect is negatively correlated with the text length, with longer texts ($>2,000$ characters) being almost not affected at all by the performed manipulation. Shorter texts, however, can be heavily affected. In almost all cases, adding one of the two expressions yields an increased score, which can grow up to, e.g., 0.7 for identity attack. For all three attributes, the difference is larger when adding `\#Juden' than `\#Israel'. Moreover, the effect is least pronounced for the attribute toxicity (mean: 0.04 for `\#Israel' and 0.08 for `\#Juden'), and strongest for identity attack (mean: 0.13 for `\#Israel' and 0.2 for `\#Juden'). The latter is not so surprising given the fact that both additions are strongly related to Jewish identity. 
Assuming a lower threshold of 0.5 for a text to be further analyzed by scientists or monitored by content moderators, this would imply an increase from 72\% to 92\% resp. 96\% for identity attack, a significantly smaller increase from 49\% to 57\% resp. 68\% for toxicity, and a growth from 40\% to 48\% resp. 59\% for severe toxicity.  

\begin{figure}[h!]
\begin{center}
\includegraphics[width=\linewidth]{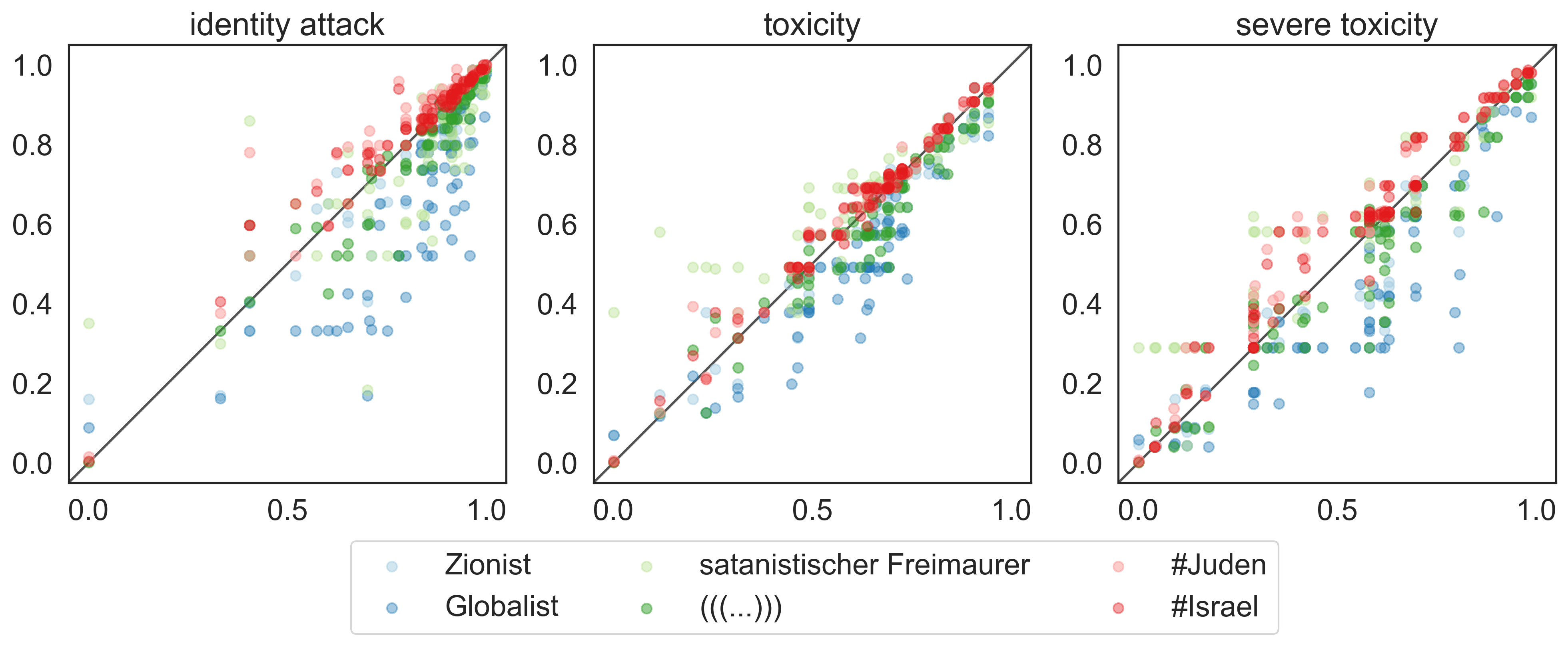}%
\end{center}
\caption{Score of original message vs. manipulated text ($n=89$): replacing words related to Jewishness by codes decreases the scores in most cases.}
\label{fig:6}
\end{figure}

Our second experiment indicates that using antisemitic codes instead of directly mentioning Jews decreases the scores in most of the cases. Figure \ref{fig:6} shows that the code `Globalist' has the strongest decreasing effect, making it an attractive term for users interested in disseminating antisemitic content without being moderated or banned from a discussion. Using the code `Zionist' or adding triple parentheses has some decreasing effects as well, though not as strong as `Globalist'. In some cases, we observed that using codes actually had an increasing effect on the respective scores. This was mainly the case for the code `satanistischer Freimauer'. We assume that this is due to the negative connotations of the adjective `satanistisch'. The effects of manipulations depend on the score of the original message, with those already assessed with a score near 1 being least effected. Interestingly, toxicity is least affected by the manipulations.

Considering again 0.5 as a threshold for all three attributes, the codes `globalist' and `zionist' significantly reduce the number of detected messages, while the usage of triple parentheses has a rather small reducing effect, and satanist Freemason as well as appending \#Israel or \#Juden yield a clear increase (cf. ~Table \ref{tab:2}). 

\begin{table}[]
\begin{tabular}{p{0.3\linewidth} | p{0.18\linewidth} | p{0.18\linewidth} | p{0.18\linewidth} }
 & identity attack & toxicity & severe toxicity \\
\hline
original text     & { 85} & {69} & {60} \\
Zionist           & { 84} & {59} & {50} \\
Globalist        & { 73} & {49} & {38} \\
satanistischer Freimaurer & { 86} &  {76} & {67} \\
(((...)))        & {85} & {66} & {55} \\
\#Israel        & { 87} & {72} & {64}\\
\#Juden          & {87}  & {73}  & {67}
\end{tabular}
\caption{Number of explicitly antisemitic texts with affirmative stance with scores above $0.5$ ($n=89$) after applying different manipulations.}
\label{tab:2}
\end{table}

\section{Summary}
We analyzed Perspective API's assessment of the toxicity of antisemitic online content using a German-language Telegram and Twitter dataset. We  conducted two experiments to examine the sensitivity of the API towards the direct mention of Jews and Israel, compared to cases in which these terms were replaced by different codes. 

Regarding RQ1, texts with antisemitic content were generally scored as more toxic than texts without such content, with median scores approximately two times higher. This indicates that texts with antisemitic content share relevant features with texts rated as toxic by the API. We observed the greatest differences between the positive and negative class regarding identity attack and severe toxicity. 

Explicit forms of antisemitism are rated as more toxic, more threatening, etc. than forms of encoded antisemitism. In combination with our findings for RQ1, this indicates that Perspective API generally does recognize antisemitic content as toxic, but finds it more difficult to recognize rather implicit forms of antisemitism. 

Texts taking a critical stance towards antisemitism are rated with higher scores compared to both texts with a neutral/uncertain and an affirmative stance, the latter two being  similarly rated. As demonstrated with two qualitative examples, a statement clearly expressing Holocaust denial is not rated as toxic, while a statement clearly critical of the Holocaust receives very high scores. With respect to RQ3, our findings demonstrate that Perspective API is not able to differentiate the stance of texts even at a basic level.

To assess the effect of modifying antisemitic statements, we performed two experiments. Adding direct mentions of Jews or Israel to the end of a text resulted in an increase of scores, particularly for shorter posts, indicating that Perspective API is sensitive to the use of identity-group words without necessarily taking into account their textual context. In contrast, replacing direct mentions of Jews resulted in a decrease of scores in most of the cases. We observed the strongest decreasing effect for the code `Globalist', followed by `Zionist' and the use of triple parentheses around words like e.g. `Jude' (Jew) or `jüdisch' (Jewish). In some cases, the use of codes actually resulted in an increase of scores. This was mostly the case for the code `satanistischer Freimaurer' (satanic Freemason), probably due to the negative connotation of the adjective `satanisch'. 

\section{Discussion}
Perspective API is already widely used in moderation and monitoring of comments but also as a tool in the study of online communication, including antisemitic speech.  
Our findings indicate that on a basic level, Perspective API recognizes antisemitic content as toxic.
When taking a closer look, however, our investigation reveals several limitations and critical weaknesses of the service, both for research and content moderation tasks. 
While it reacts to explicit forms of antisemitism, it will most likely miss rather subtle and implicit forms.
According to our results, a lot of texts classified as antisemitic would be neglected by research projects with a toxicity threshold of 0.8 for 
data collection. 
Even with a lower threshold of 0.5, more than half of the texts expressing encoded or post-Holocaust antisemitism, and around a third of explicitly antisemitic texts in our corpus, would not be considered.
This indicates that Perspective API is able to detect only the most blatant manifestations of antisemitism. This is a severe limitation of the API considering the implicitness of antisemitism, its often encoded character, but also regarding forms such as post-Holocaust antisemitism which primarily function via self-victimization instead of direct attacks against Jewish individuals or communities. 

Furthermore, the API clearly struggles with correct stance interpretation.
This is a critical weakness, for using the service to build research corpora and even more for the task of content moderation. 
Our results indicate that Perspective API is more sensitive to (potentially harsh) critiques of antisemitism rather than to affirmative antisemitic statements. This calls for further critical research and evaluation, also with regard to the impact of this bias e.g. for content moderation in social media, since such a bias penalizes counter-speech and critical discourse about antisemitism. 
 
Last but not least, our adversarial attacks against the API have demonstrated that even simple text manipulations can noticeably influence the scores. On the one hand, the service showed a bias towards the presence of identity-related keywords such as `Juden' or `Israel', assigning higher scores to texts where these words were added (cf. \cite{jigsaw_unintended_2021,rottger_hatecheck_2021}. This bias can negatively affect content moderation processes since it skews the focus towards identity-related phrases independent of their context.

On the other hand, it takes only simple manipulations in order to noticeably decrease the assigned scores. This makes it rather easy to bypass content moderation based on the API's results by using simple and known antisemitic codes, e.g. replacing terms like Jew with `Globalist'. This facilitates the inconspicuous expression and dissemination of antisemitism and undermine monitoring efforts as well as moderation policies - a problem which should not be underestimated regarding the strategical behaviour of users on online platforms to circumvent regulation and moderation policies assisted by machine learning technologies. \citet{weimann_digital_2020} have analyzed the `new language' of Right-wing extremists, a language which has partly emerged as a direct counter-reaction to the research initiative behind Perspective API. 
``To prevent violating the abuse policies of social media platforms and also to avoid detection by automatic systems like Google's Conversation AI, Far-right extremists have begun to use code words (a movement termed Operation Google) and thus a new type of hateful online language appears to be emerging: The systematic use of innocuous words to stand in for offensive racial slurs.'' 
\cite{weimann_digital_2020}

This clearly shows that human language is not only dynamic, it is also used strategically, especially in contested spaces of social and political communication. Both aspects make it difficult for automated tools to meet their claims of making those spaces less toxic, and more inclusive. So when using these tools, be it for research, monitoring, or moderation, we need to be fully aware of the limitations these tools bring along.

When using the API for research purposes, whether for sample selection or for analyzing the toxicity of online narratives, we believe that researchers should not rely solely on the automated assessment provided by the API. Rather, we recommend a thorough manual review of the obtained results before processing them further. 
The same applies to its use for content moderation. However, evaluating Perspective's results itself can be a challenging task. We believe that this requires continuous training for e.g. content moderators to enable them to recognize antisemitism in its various shapes. 
Existing annotation efforts regarding antisemitic online content consistently show that classifying antisemitism is a complex and challenging task that depends on many factors such as the context of a message or the background knowledge of the annotators \cite{ozalp_antisemitism_2020,jikeli_toward_2022}. The constant evolvement of new antisemitic codes adds to this complexity. A profound and constantly updated knowledge of these codes is thus of crucial importance. 
Last but not least, content moderation should be aware of the API's potential hypersensitivity towards certain keywords, and to expressions of counter-speech. Relying on API scoring might result in unintended punishment of counter-speech, providing another reason for continuous manual sample checks of results. The hypersensitivity towards keywords might result in unintended regulation of educational content, as incidents on a number of social platforms have shown \cite{sales_2021}. Furthermore, easily accessible feedback procedures for users should be provided to allow for early error correction and monitoring. Last but not least, transparency of the algorithms and tools used for automated detection needs to be increased, to raise awareness of the potentials and limitations of such tools, and to foster research into their strengths and weaknesses so that they can be improved. 

\bibliographystyle{acl_natbib}  
\bibliography{bibliography}
\appendix
\pagebreak
\section{Appendix}
\begin{table*}[h]
\begin{tabular}{p{0.15\textwidth} | p{0.75\textwidth} }
\textbf{Score type}      & \textbf{API Definition}                                                                                                                                                                                                                                                                          \\
\hline
insult          & Insulting, inflammatory, or negative comment towards a person or a group of people.                                                                                                                                                                                                     \\
identity attack & Negative or hateful comments targeting someone because of their identity.                                                                                                                                                                                                               \\
threat          & Describes an intention to inflict pain, injury, or violence against an individual or group.                                                                                                                                                                                             \\
toxicity        & A rude, disrespectful, or unreasonable comment that is likely to make people leave a discussion.                                                                                                                                                                                        \\
severe toxicity & A very hateful, aggressive, disrespectful comment or otherwise very likely to make a user leave a discussion or give up on sharing their perspective. This attribute is much less sensitive to more mild forms of toxicity, such as comments that include positive uses of curse words.
\end{tabular}
\caption{Attributes from Perspective API that, based on the provided definitions, should react with higher scores towards texts with antisemitic content.}
\label{tab:appendix:1}
\end{table*}
\begin{table*}[h]
\begin{tabular}{p{0.14\textwidth} | p{0.18\textwidth} | p{0.18\textwidth} | p{0.19\textwidth} | p{0.21\textwidth} }
\textbf{original word} & \textbf{triple parentheses} & \textbf{code: Zionist} & \textbf{code: Globalist}  & \textbf{code: satanistischer Freimaurer}                                    \\
\hline
Jude & (((Jude)))  & Zionist & Globalist & satanistischer Freimaurer  \\
JÜDISCH!  & (((JÜDISCH!)))  & ZIONISTISCH!  & GLOBALISTISCH!  & SATANISTISCH! \\
Antifajuden             & (((Antifajuden)))   & Antifazionisten & Antifaglobalisten  & satanistische  Antifafreimaurer                                             \\
Judenschwein  & (((Judenschwein))) & Zionistenschwein & Globalistenschwein & satanistisches Freimaurerschwein \\
\end{tabular}
\caption{Examples illustrating the modifications performed in texts labeled as explicitly antisemitic and containing words referring to Jews and Jewishness.}
\label{tab:appendix:2}
\end{table*}

\end{document}